\documentclass[10pt,twocolumn,letterpaper]{article}

\usepackage{iccv}
\usepackage{times}
\usepackage{epsfig}
\usepackage{graphicx}
\usepackage{amsmath}
\usepackage{amssymb}
\usepackage{booktabs}
\usepackage{multirow}

\usepackage{graphicx}
\usepackage{float}
\usepackage{subfig}
\usepackage{array}
\usepackage[accsupp]{axessibility}  

\usepackage[pagebackref=true,breaklinks=true,letterpaper=true,colorlinks,bookmarks=false]{hyperref}

\iccvfinalcopy 

\def\iccvPaperID{7153} 

\ificcvfinal\fi

\begin{document}

\title{Event-Guided Procedure Planning \\ from Instructional Videos with Text Supervision}

\author{An-Lan Wang$^{*}$, Kun-Yu Lin$^{*}$, Jia-Run Du, Jingke Meng$^{{\dagger}}$, Wei-Shi Zheng$^{{\dagger}}$\\
\normalsize School of Computer Science and Engineering, Sun Yat-sen University, China\\
\normalsize Key Laboratory of Machine Intelligence and Advanced Computing, Ministry of Education, China\\
{\tt\small \{wanganlan, linky5, dujr6\}@mail2.sysu.edu.cn, mengjke@gmail.com, wszheng@ieee.org}
}

\maketitle
\ificcvfinal\thispagestyle{empty}\fi

\footnotetext{ * indicates equal contribution. $\dagger$ indicates the corresponding author.}
\begin{abstract}
In this work, we focus on the task of procedure planning from instructional videos with text supervision, where a model aims to predict an action sequence to transform the initial visual state into the goal visual state. 
A critical challenge of this task is the large semantic gap between observed visual states and unobserved intermediate actions, which is ignored by previous works. 
Specifically, this semantic gap refers to that the contents in the observed visual states are semantically different from the elements of some action text labels in a procedure. 
To bridge this semantic gap, we propose a novel event-guided paradigm, which first infers events from the observed states and then plans out actions based on both the states and predicted events. 
Our inspiration comes from that planning a procedure from an instructional video is to complete a specific event and a specific event usually involves specific actions. 
Based on the proposed paradigm, we contribute an Event-guided Prompting-based Procedure Planning (E3P) model, which encodes event information into the sequential modeling process to support procedure planning. 
To further consider the strong action associations within each event, our E3P adopts a mask-and-predict approach for relation mining, incorporating a probabilistic masking scheme for regularization. 
Extensive experiments on three datasets demonstrate the effectiveness of our proposed model.
\end{abstract}

\section{Introduction}

In this work, we focus on the procedure planning task from instruction videos~\cite{chang2020procedure,2021EXTGAIL,sun2022plate,zhao2022p3iv}. 
Given the current state (a frame or a clip), procedure planning aims to predict a sequence of actions to reach a desired goal state. 
This goal-driven decision-making capability comes naturally to humans but is difficult for machine learning systems to acquire.
Therefore, due to its wide real-world applications, \eg, Autopilot~\cite{wulfmeier2016autopilot} and Robotic systems~\cite{andrychowicz2020Robotic}, solving procedure planning is of great significance.

\begin{figure}
    \centering
    \includegraphics[width=0.45\textwidth]{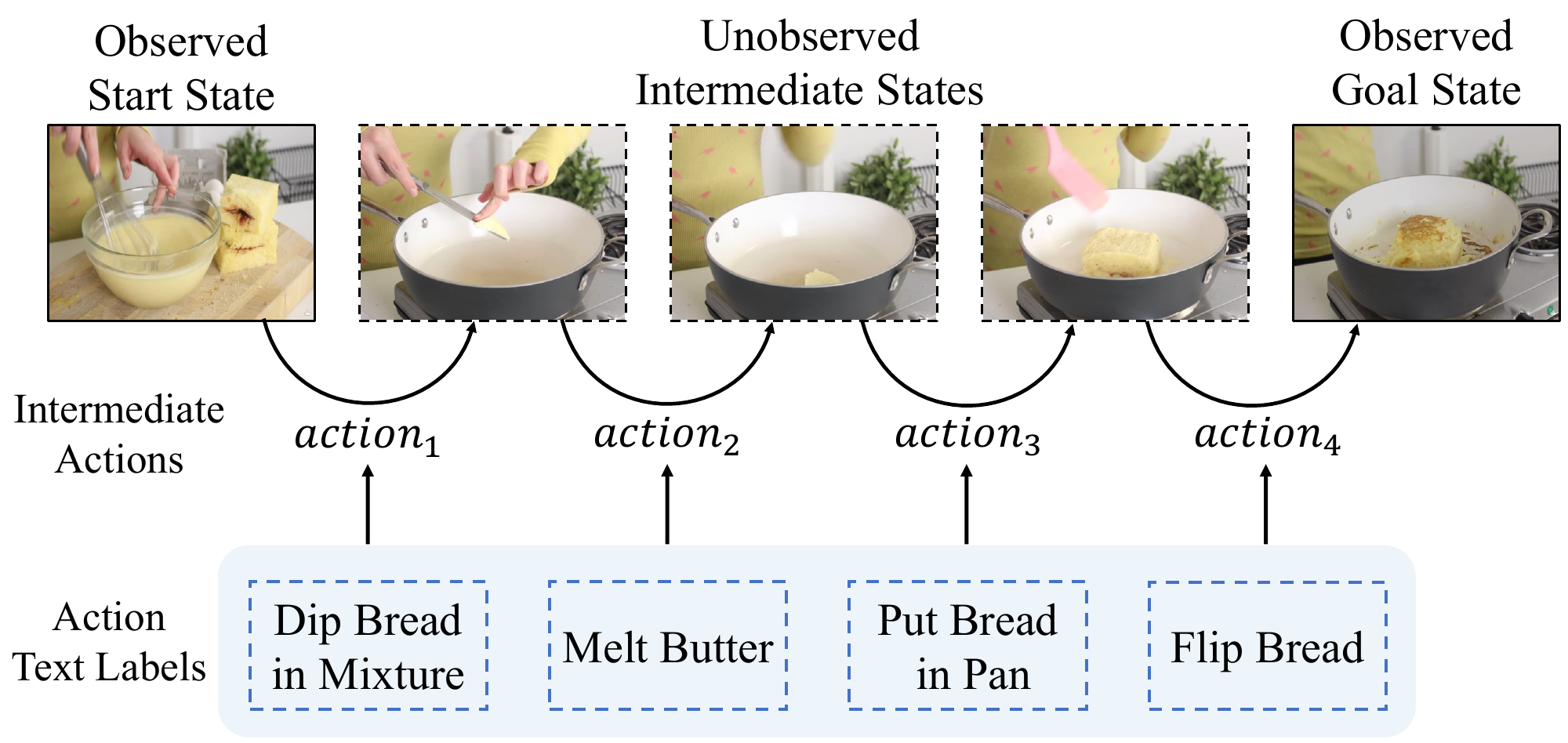}
    \caption{
    Illustration of the \textit{semantic gap}, \ie, the contents in the observed visual states are semantically different from the elements of some action text labels in a procedure. We show a four-action procedure as an example. As shown above, it is difficult to predict that we should ``Melt Butter'' by observing the start and goal states, since we see no butter but some other things (\eg, bread, pan, mixture) from the observed states. However, if we take the event of procedure ``Make French Toast'' into thinking, ``Melt Butter'' is an indispensable step. Best viewed in color.
    }
    \label{fig:motivation}
\end{figure}

\par 
Early works~\cite{chang2020procedure,2021EXTGAIL,sun2022plate} typically address the procedure planning task in an auto-regressive manner, following traditional sequential modeling works~\cite{graves2012LSTM,medsker2001RNN,chung2014GRN,2017Attention}. Specifically, given both intermediate action labels and intermediate visual states as supervision, these works adopt two-branch networks to predict the action labels and representations of states separately, based on the input start and goal states. These methods mainly differ in the feature extractor for sequential modeling, \eg, DDN~\cite{chang2020procedure} uses RNNs~\cite{medsker2001RNN}, Plate~\cite{sun2022plate} uses Transformers~\cite{vaswani2017attentionisallyouneed}. However, all these methods need access to intermediate visual states for supervision. In such a setting, it is necessary to precisely identify the start and end timestamps of all actions in training videos, which is time-consuming and labor-intensive for annotation. 

\par

\par
Recent work~\cite{zhao2022p3iv} provides a way to reduce annotation efforts. They study a weakly-supervised setting that removes the need for intermediate visual states as supervision, named Procedure Planning from instructional videos with Text Supervision (PPTS). In PPTS, text representations of intermediate action labels are introduced for supervision, leveraging the power of a pre-trained vision-language model~\cite{miech2020s3d}. To tackle PPTS, P3IV~\cite{zhao2022p3iv} proposes a memory-augmented Transformer for sequential modeling. 

In previous PPTS methods, the prediction of intermediate actions is conditioned on only the observed start and goal visual states. However, it is challenging to build a \textit{direct} connection between \textit{observed} visual states and \textit{unobserved} intermediate actions, due to a large semantic gap between them. This semantic gap refers to that the contents in the observed visual states are semantically different from the elements of some action text labels in a procedure. For example, as shown in Figure~\ref{fig:motivation}, it is difficult to directly predict the action \textbf{``Melt Butter''} by observing the start and goal states, since we see no butter but some other things (\eg, bread, pot, mixture) from the observed states.

\par
\begin{figure}
    \centering
    \includegraphics[width=0.47\textwidth]{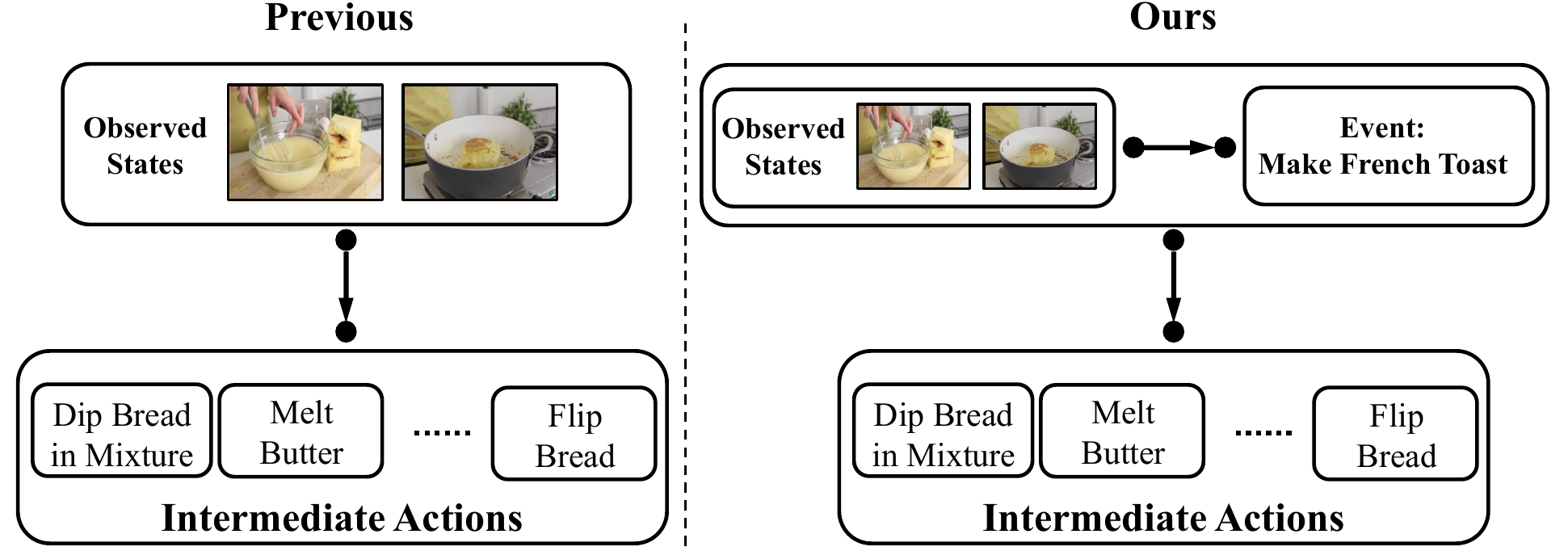}
    \caption{
    Previous methods build a \textit{direct} connection between the \textit{observed} start and goal visual states and \textit{unobserved} intermediate actions, ignoring a large semantic gap between them. In contrast, our work proposes a novel event-guided paradigm to bridge the semantic gap.} 
    \vspace{-3mm}
    \label{fig:paradigm}
\end{figure}

To bridge the semantic gap, we propose a novel event-guided paradigm (as shown in Figure~\ref{fig:paradigm}), which is not explored by previous works. 
Our proposed event-guided paradigm first infers the events of procedures based on the observed visual states and then predicts a sequence of actions based on both the states and predicted events. 
Our inspiration comes from the fact that planning a procedure from an instructional video is to complete a specific event (\ie, a procedure matches a clear intention). 
And, since a specific event usually involves specific actions, we can use the event information to support the procedure planning. 
For example, as shown in Figure~\ref{fig:motivation}, after identifying the event \textbf{``Make French Toast''} from the observed visual states, we can plan out the action \textbf{``Melt Butter''}, since melting butter is essential to attain crispy French toast. 
In addition, there are usually strong associations between actions within an event, which can be utilized for planning a reasonable procedure. 
Also shown in Figure~\ref{fig:motivation}, suppose we already know that this procedure is to make French toast and the first three actions are \textbf{``Dip Bread in Mixture$\rightarrow$ Melt Butter$\rightarrow$ Put Bread in Pan''}, we can deduce that the fourth action should be \textbf{``Flip Bread''} because no one will make French toast with only one side fried.

\par
We contribute an Event-guided Prompting-based Procedure Planning (E3P) model based on our proposed event-guided paradigm. Given event labels as supervision, our proposed E3P uses an Event-aware Prompt Generator to encode event information into the hand-crafted prompts of intermediate actions. We find that the events can generally be inferred from the observed start and goal visual states. 
After sequential modeling based on event-aware prompts, we propose an Action Relation Mining module to model the associations between actions within each event. Our Action Relation Mining module adopts a mask-and-predict approach and incorporates a probabilistic masking scheme for regularization, aiming to fully consider the action associations during training. We conduct extensive experiments on three datasets, and the results demonstrate that our proposed E3P outperforms previous state-of-the-art methods by a large margin.

\begin{figure*}
    \centering
    \includegraphics[width=1\linewidth]{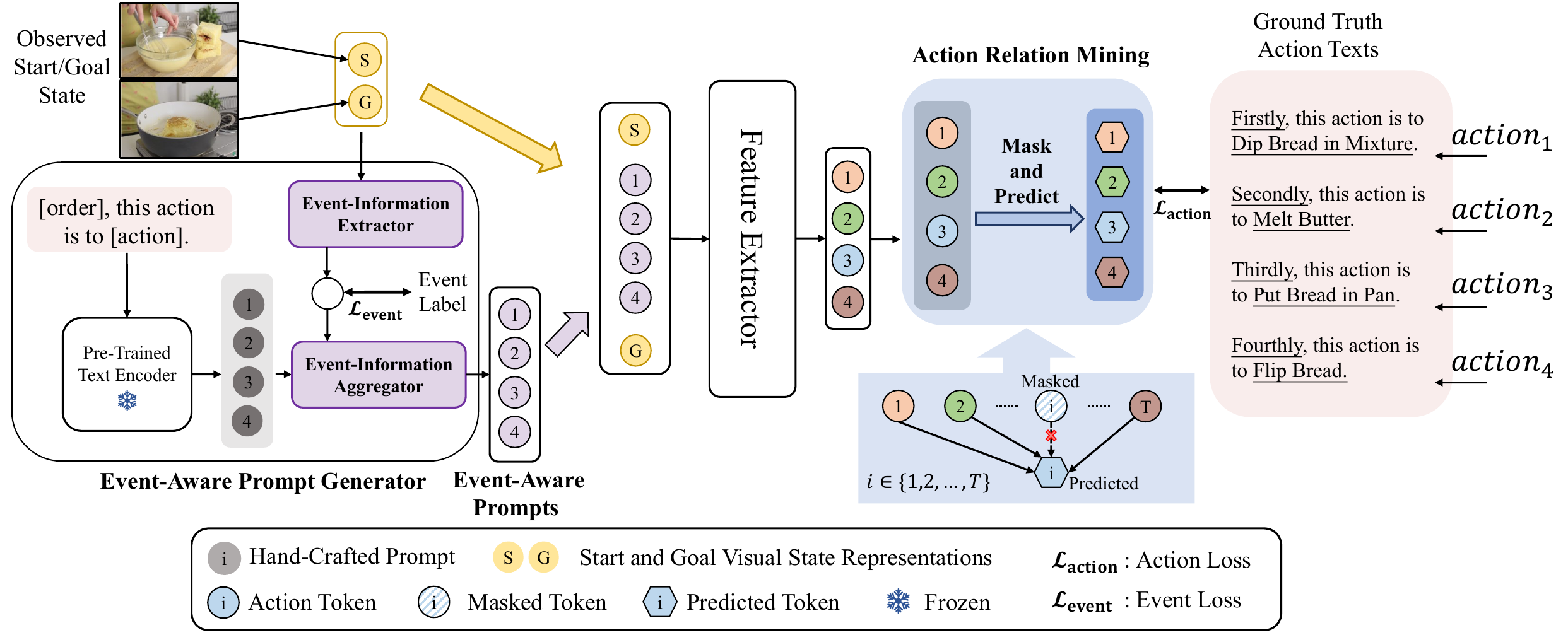}
    \caption{
    Overview of our Event-guided Prompting-based Procedure Planning (E3P) model. 
    Our proposed E3P model follows a novel event-guided paradigm to bridge the semantic gap between observed visual states and unobserved intermediate actions. 
    In this figure, we take a four-action procedure as an example.
    First of all, we use a pre-trained text encoder to extract the representations of hand-crafted action prompts. 
    Supervised by event labels, we extract event information from the given visual states and integrate them into the prompts to generate Event-Aware Prompts. 
    After sequential modeling by the feature extractor, the Action Relation Mining module exploits the action associations adopting a ``mask-and-predict'' approach. 
    Best viewed in color.
    }
    \label{fig:Pipeline}
\end{figure*}

\section{Related Works}
\subsection{Procedure Planning}
Procedure planning from instructional videos aims to predict a reasonable plan conditioned on the start and goal visual states. 
Early works on procedure planning adopt a two-branch auto-regressive method (\ie, action and visual branch) to predict actions and visual representation of intermediate states.
These works involve different network architectures for modeling, varying from recurrent neural networks~\cite{chang2020procedure}, transformers~\cite{sun2022plate} to adversarial networks~\cite{2021EXTGAIL}. 
Recently, Zhao~\etal~\cite{zhao2022p3iv} proposes Procedure Planning from instructional videos with Text Supervision (PPTS), where text representations of intermediate actions are introduced as supervision. 
To address PPTS, P3IV~\cite{zhao2022p3iv} proposes a memory-augmented Transformer for sequential modeling.
Different from previous works, we propose a novel event-guided paradigm for PPTS, aiming to bridge the semantic gap between observed visual states and unobserved intermediate actions.

\subsection{Action Recognition} 
With the success of deep learning, effective video classification architectures for action recognition have been proposed, including RNNs~\cite{donahue2015RnnsActionRecognition1,yue2015RnnsActionRecognition2,ullah2017RnnsActionRecognition3},
2D CNNs~\cite{he2016deep2Dconv3, lin2019deep2Dconv4, shao2020deep2Dconv5, simonyan2014tsn}
and 3D CNNs~\cite{miech2020s3d,carreira2017deep3D1,tran2015deep3D2,xie2017deep3D3}. 
Recently, with the success of Vision Transformer~\cite{dosovitskiy2020VIT1,liu2021VIT2}, many works adopted Vision Transformer for action recognition~\cite{girdhar2019videoTransformer1,neimark2021videoTransformer2,zhang2021VideoTransformer3,zhang2021videoTransformer4}. 
Different from the traditional action recognition task, the procedure planning task studied in our work is much more challenging, since the relation between actions should be taken into account beyond predicting individual actions. 

\subsection{Action Anticipation} 
Different from procedure planning based on start and goal states, action anticipation aims to predict future actions based on past states. 
Early action anticipation works~\cite{kataoka2016shorttermanticipation1,vondrick2016shorttermanticipation2} focus on predicting a single future action within a few moments.
Farha~\etal~\cite{abu2018will} proposed long-term action anticipation to predict a sequence of future actions by using two networks (\ie, a RNN and a CNN). 
To reduce the error accumulation caused by iterative predictions, Some methods~\cite{2019UAAA,abu2018will,ke2019time} took action labels of past states as input. 
Gong~\etal~\cite{gong2022future} adopted an end-to-end transformer model to anticipate all future actions in parallel. 
Procedure planning is similar to an action anticipation task with goal guidance, leading to more constraints for modeling. 

\subsection{Prompting and Regularization} 
There have been longstanding efforts for prompt design in NLP~\cite{liu2023pre,schick2020exploiting}. 
Recently, CLIP~\cite{radford2021CLIP} explored prompting for image understanding by formulating a image-text matching problem. 
Some other works~\cite{ju2022prompting,nag2022zero,wang2021actionclip,li2022bridge} explored prompting for video understanding. 

To alleviate the overfitting problem in deep neural networks, many dropout-like techniques were proposed for regularization~\cite{hinton2012dropout,wan2013dropconnect,ba2013adaptiveDrop,wang2013fastDrop}.
Specific to procedure planning, our proposed Action Relation Mining module involves a probabilistic masking scheme for regularization, aiming to fully consider the action associations. 

\section{Event-Guided Procedure Planning}
In this section, we elaborate on our proposed Event-guided Prompting-based Procedure Planning (E3P) model, which follows a novel event-guided paradigm.

\subsection{Problem Formulation and Model Overview}
\textbf{Problem Formulation.}
Our task is procedure planning from instructional videos with text supervision. 
Given the start visual state $o_s$ and goal visual state $o_g$, a model aims to predict a procedure of $T$ action steps, \ie, an action sequence $\{a_1,...,a_T\}$, transforming the visual state from the $o_s$ to $o_g$. 
The number of actions $T$ is provided, also known as the prediction horizon.
Following Zhao~\etal~\cite{zhao2022p3iv}, we use text representations of action labels as supervision for training.

\textbf{Model Overview.}
Figure~\ref{fig:Pipeline} illustrates the proposed E3P model based on prompting-based feature modeling. 
Given the state representations $o_s$ and $o_g$, we extract the event information of the procedure through an event-information extractor and then encode it into hand-crafted text prompts to generate Event-Aware Prompts. 
Subsequently, the Event-Aware Prompts and visual state representations are then fed into the feature extractor for sequential modeling, producing $T$ action tokens. 
These tokens are then fed into an Action Relation Mining module, which uses a ``mask-and-predict'' approach to model the relation between actions. Finally, we output a sequence of actions, namely, the procedure. Notably, we do not directly predict the distribution over possible actions, and instead by first predicting the feature representations of actions and then predicting the distribution over actions (i.e., by calculating the similarity between predicted action features and all action text features).

\subsection{Prompting-based Feature Modeling}
\label{method-ordinal-prompt}
First of all, we introduce the prompting-based feature modeling of our approach. 
Since the PPTS task requires predicting a procedure (\ie, an action sequence), and this sequence is order-sensitive, we leverage the power of a pre-trained vision-language model.
Specifically, we use a hand-crafted text prompt in the format of \textit{``[order], this action is to [action]''} 
as input, which contains an order blank and an action blank. 
Then, we obtain the representations of prompts using a pre-trained vision-language model.

We use the representations of ground truth action texts as supervision. 
In specific, we construct a sentence in the same format as the above hand-crafted prompts, with the two blanks filled in. 
For example, if the first action is ``Melt Butter'', the constructed sentence would be ``Firstly, this action is to Melt Butter''. 
Then we use the vision-language model to encode it into text representations. 
Following P3IV~\cite{zhao2022p3iv}, we use the same loss function~\cite{gutmann2010contrastive} to supervise the model, which is formulated as follows:
\begin{equation}
    \mathcal{L}_{action}=-\sum_{t=1}^{T}\left[\log \frac{\exp \left(l_{+} \cdot \widetilde{a}_{t}\right)}{\sum_{j=1}^N \exp \left(l_{j} \cdot \widetilde{a}_{t}\right)}\right],
\end{equation}
where $l_+$ is the ground truth action text presentation, $l_j$ is the text representation of the $j$-th action, $N$ is the number of actions in the dataset and $\widetilde{a}_{t}$ is the $t$-th action token (the final output of our model).

\subsection{Event-aware Prompt Generator}
\label{method-intention-aware}
To bridge the semantic gap between the observed visual states and unobserved intermediate actions, we propose an Event-aware Prompt Generator that encodes the event information to guide the procedure planning process.
Our inspiration comes from that planning a procedure from an instructional video is to complete a specific event and a specific event usually involves specific actions.
Our Event-aware Prompt Generator mainly contains two parts, \ie, an event-information extractor and an event-information aggregator, aiming to extract event information and encode event information into the text prompts, respectively.
 
Specifically, the event-information extractor $\mathbf{E_{e}}$ extracts the event information $\hat{e}$ from the start state $o_s$ and goal state $o_g$, which is given as follows: 
\begin{equation}
    \hat{e} = \mathbf{E}_{e}(o_s, o_g). 
\end{equation}
To guide the event information extraction, we stack a classification head $h_e(\cdot)$ on top of the $\mathbf{E_{e}}(\cdot, \cdot)$. 
And, the event loss is given as follows:
\begin{equation}
    \mathcal{L}_{event}= \mathrm{CE}(h_e(\hat{e}), y_e), 
\end{equation}
where $\mathrm{CE}$ is the cross-entropy loss and $y_e$ is the ground truth event label provided by the dataset.

Then, we encode the event information $\hat{e}$ into the hand-crafted prompt representations $p_{1:T}$ using the event information aggregator.
The event-information aggregator takes $T$ prompt representations and the event information as inputs and produces $T$ event-aware prompts, which is formulated as follows:
 \begin{equation}
    p_{1:T}^{e} = \mathcal{F}(p_{1:T}, \hat{e}),
\end{equation}
where  $\mathcal{F}(\cdot)$ is the event-information aggregator and $p_{i:T}^{e}$ are the event-aware prompts. 
The introduction of event information would constrain our model to predict actions more related to the event of procedure.

Next, we concatenate generated event-aware prompts with start and goal visual state representations. 
After positional encoding, we input the event-aware prompts and visual states into the feature extractor for sequential modeling. 
The input of the feature extractor is given as follows: 
\begin{equation}
    Q=[o_s, p_{1}^{e},p_{2}^{e},...,p_{T}^{e},o_g],
\end{equation}
where $p_{i}^{e}$ is the $i$-th event-aware prompts. 
After sequential modeling, the feature extractor outputs $T+2$ tokens. We take the middle $T$ action tokens $\hat{a}_{1:T}$ as input to our next module.

\subsection{Action Relation Mining within Events}
\label{method-action-reasoning}
In this section, we propose to model the relation between actions within individual events to support the procedure planning. 
Usually, there are  strong associations between actions within an event, which can be utilized for planning a reasonable procedure. 
Accordingly, we propose an Action Relation Mining (ARM) module exploiting a ``mask-and-predict'' approach, which refines the prediction of procedure planning by mining the relation between actions within events. 

For our ARM module, we adopt the masked self-attention with a specially designed mask as core. 
Specifically, the input to the ARM module is a list of action tokens, 
\ie, $\hat{Q} = [\hat{a}_{1}, \hat{a}_{i},..., \hat{a}_{T}]$. 
For the masked self-attention, we use a \textit{deterministic} mask $M\in\mathbb{R}^{T \times T}$, where all elements on the main diagonal are manually set to zero. 
Such a mask design ensures that our ARM model can leverage the information of all other actions for the prediction of the one masked action.
The \textit{deterministic} mask is defined as:
\begin{equation}
    \begin{aligned}
    M_{i,j}=
	&\left\{\begin{aligned}
        &0,\quad \text{if}\ i=j, \\
        &1,\quad \text{otherwise,} \\
	\end{aligned}\right.\\
	\end{aligned}
\end{equation}
where $i$ and $j$ represent the row and column of the attention mask matrix respectively. 
After feature modeling by our ARM module, we obtain the final prediction of action sequence by a residual connection, \ie. $\widetilde{a}_{1:T} = \hat{a}_{1:T} + \check{a}_{1:T}$, 
where $\check{a}_{1:T}$ is the prediction of ARM module and $\widetilde{a}_{1:T}$ is the final prediction. 
Intuitively, our ARM module conducts a refinement on the basic procedure prediction $\hat{a}_{1:T}$. 

\begin{figure}[t]
    \centering
    \vspace{-3mm}    \includegraphics[width=0.45\textwidth]{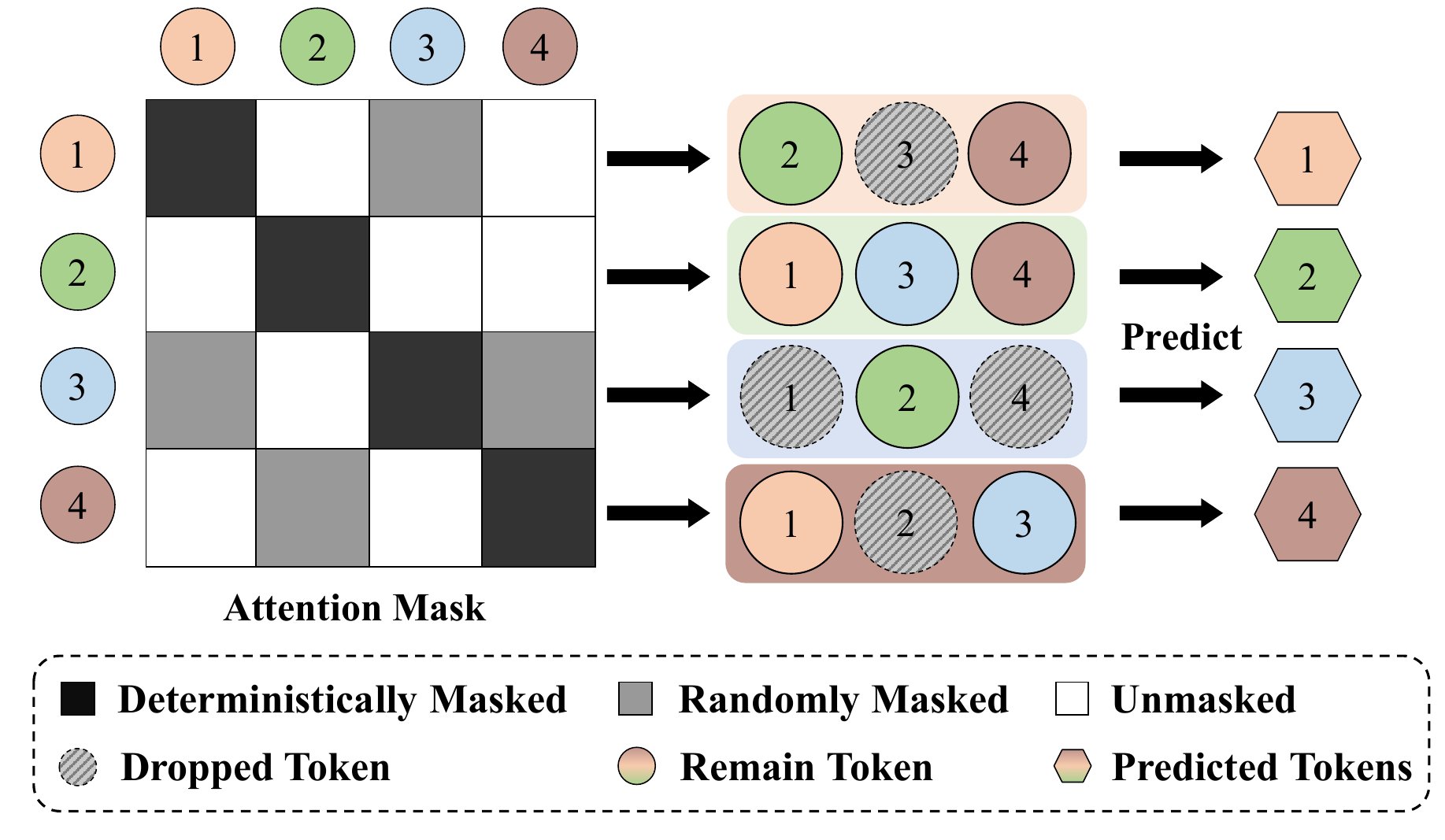}
    \vspace{-1mm}
    \caption{
    Illustration of the mask in our Action Relation Modeling module.
    We take a four-action procedure as an example. 
    For the attention mask, we first set the elements of the main diagonal to zero and then randomly set some other elements to zero by a drop rate~$\tau$.}
    \vspace{-4mm}
    \label{fig:action-reasoning}
\end{figure}

\textbf{DropRelation.}
In the above relation mining process, we adopt a ``mask-and-predict'' approach with only one token masked out.
However, such a modeling focuses on modeling the relation between the one masked action token and all other action tokens (\eg, one masked action token and three unmasked action tokens in a four-action procedure) during training, thus it may miss some associations between actions. 
For example, suppose we have a four-action sequence ``Action 1 $\rightarrow$ Melt Butter $\rightarrow$ Action 3 $\rightarrow$ Flip Bread'', where the first and third actions are unknown. 
We are still able to infer that the third action is ``Put Bread in Pan'' according to the known two actions, without knowing the first action. 
Therefore, to fully consider the action associations, we equip our ARM module with a regularization technique named DropRelation. 

Specifically, in addition to dropping tokens in the main diagonal, we randomly drop some of the other tokens during the ``mask-and-predict'' process. 
For each item in the mask, there is a random variable following a uniform distribution, denoted by $\alpha_{i,j}$. 
According to the $\alpha_{i,j}$, we obtain a  \textit{probabilistic} mask $\widetilde{M}$ as follows: 
\begin{equation}
    \begin{aligned}
  \widetilde{M}_{i,j}=
		&\left\{\begin{aligned}
        1,  & \quad \text{ if } i\ne j\ \text{and} \ \alpha_{ij} >  \tau, \\
        0, & \quad \text{otherwise}, \\
	\end{aligned}\right.\\
	\end{aligned}
\end{equation}
where $\tau$ refers to the drop rate. 
In this way, we randomly drop some connections (relation) between action tokens to regularize the relation mining process.
In addition, we ensure that dropping at most $T-2$ tokens in a single row in the $\widetilde{M}_{i,j}$.
An illustration of the masking scheme in our ARM is shown in Figure~\ref{fig:action-reasoning}.

\subsection{Traning and Inference}
During training, the overall objective is as follows:
\begin{equation}
    \mathcal{L}=\mathcal{L}_{\text {action}} + \mathcal{L}_{\text {event}}.
\end{equation}

During inference, following previous methods~\cite{zhao2022p3iv}, our model only accesses to the start and goal visual states. 
Notably, the DropRelation regularization is only used during training and turned off during inference. 
Moreover, following previous work~\cite{zhao2022p3iv}, we adopt the Viterbi~\cite{viterbi1967viterbi} post process as well.
\section{Experiments}

\begin{table*}[htbp]
    \small
  \centering
  \vspace{-3mm}
  \caption{Comparison with the state-of-the-art methods on CrossTask for prediction horizon $T\in\{3,4\}$. SR, mAcc, and mIoU indicate Success Rate, mean Accuracy and mean Intersectin over Union, respectively. The numbers in bold-faced and in underline indicate the highest and the second-highest result, respectively. The column \textit{Supervision} indicates the type of supervision used in training,~\ie, visual state, and action text.
 $\dagger$ indicates that data augmentation is in usage during training.
  }
    \begin{tabular}{lccrcccccccc}
    \toprule
        \multirow{2}{*}{Methods}  & \multirow{2}{*}{Year} & \multirow{2}{*}{Supervision} &  & \multicolumn{3}{c}{$T=3$}  &       & \multicolumn{3}{c}{$T=4$} \\
\cline{5-7}\cline{9-11}     &    &     &  & SR$\uparrow$   & mAcc$\uparrow$ & mIoU$\uparrow$ &       & SR$\uparrow$   & mAcc$\uparrow$ & mIoU$\uparrow$ \\
    \midrule
      Random & - & -   & & \textless0.01  & 0.94  & 1.66  &       & \textless0.01 & 0.83  & 1.66 \\
      Retrieval-Based &  - &  Visual State &  & 8.05  & 23.30 & 32.06 &       & 3.95  & 22.22 & 36.97 \\
      WLTDO~\cite{ehsani2018let} & 2018 & Visual State  &  & 1.87  & 21.64 & 31.70 &     & 0.77  & 17.92 & 26.43 \\
      UAAA~\cite{2019UAAA} & 2019 & Visual State  &  & 2.15  & 20.21 & 30.87 &       & 0.98  & 19.86 & 27.09 \\
      UPN~\cite{2018UPN} & 2018 & Visual State &   & 2.89  & 24.39 & 31.56 &       & 1.19  & 21.59 & 27.85 \\
      DDN~\cite{chang2020procedure} & 2020 & Visual State  &  & 12.18 & 31.29 & 47.48 &       & 5.97  & 27.10 & 48.46 \\
      Ext-GAIL w/o Aug.~\cite{2021EXTGAIL} & 2021 &  Visual State   & & 18.01 & 43.86 & 57.16 &       & -     & -     & - \\
      Ext-GAIL~\cite{2021EXTGAIL}~$\dagger$ & 2021  & Visual State  &  & 21.27 & 49.46 & 61.70 &       & \underline{16.41} & 43.05 & 60.93 \\
      \hline
      P3IV w/o Adv.~\cite{zhao2022p3iv}& 2022  & Text  &  & 22.12 & 45.57 & 67.40 &       & - & - & - \\
       P3IV~\cite{zhao2022p3iv}& 2022 &  Text  &  & \underline{23.34} & \underline{49.96} & \underline{73.89} &       & 13.40 & \underline{44.16} & \underline{70.01} \\
      Ours &  - & Text &  & \textbf{26.40} & \textbf{53.02} & \textbf{74.05} &   & \textbf{16.49}     &  \textbf{48.00}     & \textbf{70.16}  \\
    \bottomrule
    \vspace{-6mm}
    \end{tabular}%
  \label{tab:maintable}%
\end{table*}%

We conduct experiments on three datasets and use three metrics to verify the effectiveness of our proposed Event-guided Prompting-based Procedure Planning (E3P) method.
\subsection{Datasets and Metrics}
\label{exp:datasetandmetircs}
\textbf{Datasets:}~We conduct experiments on the following three datasets:
(1) \textbf{CrossTask}~\cite{zhukov2019crosstask} contains instructional videos collected for 83 different events, which are divided into 18 primary and 65 related events. Following previous works~\cite{chang2020procedure,sun2022plate,zhao2022p3iv}, we use the primary events, containing 2750 videos with an average of 7.6 actions per video. (2)~\textbf{Narrated Instructional Videos (NIV)}~\cite{alayrac2016NIV} is a dataset collected from real-world instruction videos from the Internet. This dataset contains 150 videos of five events, with an average of 9.5 actions per video. (3)~\textbf{COIN} is a large labeled instructional video dataset, which is collected from YouTube and consists of 11827 videos related to 778 different actions and on average 3.6 actions per video. For all three datasets, following previous works~\cite{chang2020procedure,zhao2022p3iv}, we adopt 70\%/30\% to create train/test splits and use a shift window to curate the dataset into plans with different time horizons. 

\textbf{Metrics:}~We use three evaluation metrics: 
(1) \textbf{Success Rate (SR)} considers a procedure successful only if it exactly matches the ground truth. 
(2) \textbf{mean Accuracy (mAcc)} considers the match of single action between predicted and ground truth action sequences, where action matches the ground truth at the same timestamp is considered correct. 
(3) \textbf{mean Intersection over Union (mIoU)} treats the predicted and ground truth action sequences as two sets, and measures the overlap between them. Note that mIoU is agnostic to the order of actions, which is only used as an auxiliary metric (to measure whether a model predicts the correct action set for the procedure). For more details about these metrics, please refer to DDN~\cite{chang2020procedure}.

\subsection{Implementation Details}
For a fair comparison, we follow the previous approach~\cite{zhao2022p3iv} to extract the representations of start and goal visual states using the S3D network~\cite{miech2020s3d} pretrained on the HowTo100M~\cite{miech2019howto100m} dataset. 
The text encoder is adopted from the pre-trained CLIP because the text model of P3IV~\cite{zhao2022p3iv} cannot encode prompt sentences. 
We use a Transformer as feature extractor.

We use Adam~\cite{kingma2014adam} optimizer with a weight decay of 0.4 and set the learning rate as 7e-4. 
Our model is trained for 200 epochs with a batch size of 32. 
We report the average results over three random trials. 
The method is implemented in PyTorch~\cite{paszke2017pytorch}.
Please refer to the Appendix for more implementation details.

\subsection{Comparison with State-of-the-Arts}
On the CrossTask dataset, we compare our E3P with two types of methods in procedure planning, and our proposed E3P outperforms previous methods in all metrics as shown in Table \ref{tab:maintable}. 
Compared with previous text-supervised methods, our E3P obtains significant improvement, \eg, 3.06\% for $T=3$ and 3.09\% for $T=4$ in terms of Success Rate (SR). 
The results demonstrate that our E3P effectively captures the relation of actions and predicts more accurate procedures, which is attributed to our proposed event-guided paradigm. 
Our model performs slightly better than P3IV~\cite{zhao2022p3iv} in mIoU. 
This is because P3IV adopts an adversarial strategy during training and samples 1500 procedures in the inference phase to make the final prediction for each procedure, while we make only one prediction. 
By removing the adversarial strategy from P3IV (\ie, P3IV w/o Adv), our model outperforms it by 6.65\% ($T=3$) in terms of mIoU. 
In addition, we compare our E3P with methods that use intermediate visual states as supervision, 
and our E3P still outperforms all these methods. 

\begin{table}[t]
  \centering
  \caption{Comparison with the state-of-the-art methods on a large dataset COIN for prediction horizon $T \in \{3,4\}$. ${\dagger}$ indicates using the visual state as supervision. The bold-faced and underlined numbers indicate the highest and the second-highest performance, respectively.}
  \vspace{-1mm}
    \resizebox{0.47\textwidth}{!}{
        \renewcommand\arraystretch{1.06}
    \begin{tabular}{crcccrccc}
    \toprule
      \multirow{2}{*}{Methods}        & \multicolumn{3}{c}{$T=3$}  &       & \multicolumn{3}{c}{$T=4$} \\
\cline{2-4}\cline{6-8}         & SR$\uparrow$   & mAcc$\uparrow$ & mIoU$\uparrow$ &       & SR$\uparrow$   & mAcc$\uparrow$ & mIoU$\uparrow$ \\
    \midrule
    Random &     \textless0.01 & \textless0.01 & 2.47  &       & \textless0.01 & \textless0.01 & 2.32 \\
    Retrieval-Based${\dagger}$ &     4.38  & 17.40  & 32.06 &       & 2.71  & 14.29 & 36.97 \\
    DDN$~\cite{chang2020procedure}^{\dagger}$   &     13.90  & 20.19 & 64.78 &       & 11.13 & 17.71 & 68.06 \\
    P3IV~\cite{zhao2022p3iv}  &     \underline{15.40}  & \underline{21.67} & \underline{76.31} &       & \underline{11.32} & \underline{18.85} & \underline{70.53} \\
    Ours  &   \textbf{19.57} & \textbf{31.42} & \textbf{84.95} &       & \textbf{13.59} & \textbf{26.72} & \textbf{84.72} \\
    \bottomrule
    \end{tabular}%
    }
  \label{tab:COINdataset}%
\end{table}%

\begin{table}[t]
  \centering
  \caption{Comparison with the state-of-the-art methods on NIV dataset for prediction horizon $T \in \{3,4\}$. ${\dagger}$ indicates using the visual state as supervision. The bold-faced and underlined numbers indicate the highest and the second-highest performance, respectively.}
  \vspace{-1mm}
  \resizebox{0.47\textwidth}{!}{
    \begin{tabular}{crcccrccc}
    \toprule
       \multirow{2}{*}{Methods}  & \multicolumn{3}{c}{$T=3$}  &     & \multicolumn{3}{c}{$T=4$} \\
\cline{2-4}\cline{6-8}    & SR$\uparrow$   & mAcc$\uparrow$ & mIoU$\uparrow$ &    & SR$\uparrow$   & mAcc$\uparrow$ & mIoU$\uparrow$ \\
    \midrule
    Random   & 2.21  & 4.07  & 6.09  &       & 1.12  & 2.73  & 5.84 \\
    DDN~\cite{chang2020procedure}${\dagger}$    & 18.41 & 32.54 & 56.56 &       & 15.97 & 27.09 & 53.84 \\
    Ext-GAIL~\cite{2021EXTGAIL}${\dagger}$   & 22.11 & 42.20  & 65.93 &       & 19.91 & 36.31 & 53.84 \\
    P3IV~\cite{zhao2022p3iv}   & \underline{24.68} & \underline{49.01} & \underline{74.29} &       & \underline{20.14} & \underline{38.36} & \underline{67.29} \\
    Ours   & \textbf{26.05} & \textbf{51.24} & \textbf{75.81} &       & \textbf{21.37} & \textbf{41.96} & \textbf{74.90} \\
    \bottomrule
    \end{tabular}%
    }
  \label{tab:NIVdataset}%
  \vspace{-1mm}
\end{table}%

\begin{table}[htbp]
\scriptsize
  \centering
  \caption{Comparison with the state-of-the-art methods on CrossTask dataset for different prediction horizon $T\in\{3,4,5,6\}$. ${\dagger}$ indicates using visual states as supervision. The bold-faced and underlined numbers indicate the highest and the second-highest performance, respectively. The results are evaluated on Success Rate (SR).}
  \resizebox{0.47\textwidth}{!}{
    \begin{tabular}{lrrrr}
    \toprule
    Methods &   $T=3$    &   $T=4$    &    $T=5$   & $T=6$ \\
    \midrule
    Retrival-Based & \multicolumn{1}{c}{8.05} & \multicolumn{1}{c}{3.95} & \multicolumn{1}{c}{2.40} & \multicolumn{1}{c}{1.10} \\
    DDN~\cite{chang2020procedure}${\dagger}$   & \multicolumn{1}{c}{12.18} & \multicolumn{1}{c}{5.97} & \multicolumn{1}{c}{3.10} & \multicolumn{1}{c}{1.20} \\
    P3IV~\cite{zhao2022p3iv}  & \multicolumn{1}{c}{\underline{23.34}} & \multicolumn{1}{c}{\underline{13.40}} & \multicolumn{1}{c}{\underline{7.21}} & \multicolumn{1}{c}{\underline{4.40}} \\
    Ours  &   \multicolumn{1}{c}{\textbf{26.40}}    &    \multicolumn{1}{c}{\textbf{16.49}}   &   \multicolumn{1}{c}{\textbf{8.96}}    & \multicolumn{1}{c}{\textbf{5.76}}  \\
    \bottomrule
    \end{tabular}%
    }
  \label{tab:longhorizon}%
\end{table}%

 We also conduct experiments on the COIN and NIV datasets. The results are reported in Table~\ref{tab:COINdataset} and Table~\ref{tab:NIVdataset}. Our E3P outperforms all previous methods on both datasets. 
 On the COIN dataset, the performance of our model far exceeds the latest state-of-the-art P3IV~\cite{zhao2022p3iv} by 4.17\% ($T=3$) and 2.27\% ($T=4$) in terms of Success Rate (SR). 
 On the NIV dataset, our model improves the performance by up to 1.37\% ($T=3$) and 1.23\% ($T=4$) over the state-of-the-art method~\cite{zhao2022p3iv} in terms of SR. The consistent results on all three datasets demonstrate the effectiveness of our E3P, which attributes to our novel event-guided paradigm.

We further verify the effectiveness of our method for different prediction horizons on the CrossTask dataset. 
The results are reported in Table \ref{tab:longhorizon}, our model shows significant improvement compared with P3IV~\cite{zhao2022p3iv} in the more difficult long-time horizon prediction,~\ie, 1.75\% ($T=5$) and 1.36\% ($T=6$) in terms of Success Rate (SR).

\begin{table}[t]
    \large
    \centering
    \caption{Ablation study of our method on the CrossTask dataset for prediction horizon $T\in \{3,4\}$ in terms of Success Rate (SR) and mean Accuracy (mAcc).}
    \large
    \renewcommand\arraystretch{1.15}
    \resizebox{0.47\textwidth}{!}{
    \begin{tabular}{c|ccc|cc|cc}
    \toprule
    \multirow{2}{*}{Model} & \multirow{2}{*}{PFE}  & \multirow{2}{*}{EPG} & \multirow{2}{*}{ARM} & \multicolumn{2}{c|}{$T = 3$} & \multicolumn{2}{c}{$T = 4$} \\
     \cline{5-8} &  &  &  & SR$\uparrow$ & mAcc$\uparrow$ & SR$\uparrow$ & mAcc$\uparrow$  \\
    \hline
    baseline     &       &       &       &   22.56  & 46.17 & 12.97 & 43.97\\
    + PFE     & \checkmark     &       &       &  23.55 & 48.33 & 13.53 & 45.20 \\
    + EPG     & \checkmark     &  \checkmark    &      &  25.62 & 52.28 & 14.85 & 47.44\\
    Full     & \checkmark     & \checkmark     & \checkmark   & 26.40 & 53.02  & 16.49 & 48.00 \\
    \hline
     Full w/o EPG     & \checkmark     &      &  \checkmark     &   25.25 & 52.59 & 14.23 &47.61\\

    \bottomrule
    \end{tabular}
    }
    \label{tab:componentAblation}
    \vspace{-3mm}
\end{table}

\subsection{Ablation Study}
\label{section:ablation}
\textbf{Effect of main components in E3P.}
In Table \ref{tab:componentAblation}, we analyze the effect of each component of our proposed method. 
Following the P3IV~\cite{zhao2022p3iv}, we apply an action classifier on top of the backbone as our baseline, using action text labels as supervision.
By adopting the Prompting-based Feature Extractor (PFE), we obtain an improvement over the baseline. 
Then, by introducing Event-aware Prompt Generator (EPG), our model obtains significant improvements, \eg, 2.07\% when $T=3$ and 1.32\% when $T=4$ in terms of SR, which demonstrates the effectiveness of our proposed event-guided paradigm. 
Noteworthy, we evaluate the performance of event classification based the start and goal visual states and find that the event of procedure can generally be inferred from the observed states (\eg, the event classification accuracy is 99.5\% when $T=3$). 
Then, by introducing the Action Relation Mining (ARM) module, our model obtains further performance improvement, \eg, 0.78\% when $T=3$ and 1.64\% when $T=4$ in terms of SR, which demonstrates the effectiveness of mining action associations within each event. 
In addition, if the Event-aware Prompt Generator is removed from the  full E3P model, the performance drops but still achieves state-of-the-art. 
We mainly attribute this to the proposed ARM module, as there are strong associations between actions even without knowing the event. 
In summary, the ablation study demonstrates the effectiveness of the proposed event-guided model.

\textbf{Analysis of DropRelation Regularization.}  
In Table \ref{tab:maskrate}, we conduct a quantitative analysis of the DropRelation regularization. 
In general, for all prediction horizons (\ie, $T \in \{3,4,5\}$), the performance of our method follows a trend of first increasing and then decreasing. 
Compared with a short prediction horizon (\eg, $T=3$), the best drop rate is higher for a longer prediction horizon (\eg, $T=5$). 
This is because the longer prediction horizon requires modeling more diverse action relations, and a relatively large drop rate ensures that our Action Relation Mining module adequately covers different combinations of the action tokens during the training, thus enabling the model to adequately consider the action associations. 
A drop rate of 20\% is a recommended choice for all prediction horizons.

\begin{table}[tbp]
  \centering
  \caption{Quantitative analysis of DropRelation regularization for prediction horizon $T\in \{3,4,5\}$ on CrossTask dataset in terms of Success Rate (SR).}
  \renewcommand\arraystretch{1.12}
  \resizebox{0.47\textwidth}{!}{
    \begin{tabular}{c|cccccc}
    \toprule
    Prediction & \multicolumn{6}{c}{Drop  Rate (\%)} \\
\cline{2-7}    Horizon & 0     & 5     & 10    & 20    & 30    & 40 \\
    \hline
    $T=3$   & 26.04 & 26.40 & 26.33  & 26.26 & 25.87 & 25.76 \\
    $T=4$   & 15.69 & 15.87 & 16.18  & 16.49 & 16.12 & 15.81 \\
    $T=5$   & 7.95  & 8.23  & 8.54  & 8.77  & 8.96  & 8.68 \\
    \bottomrule
    \end{tabular}%
    }
  \label{tab:maskrate}%
\end{table}%

\begin{table}[t]
    \small
    \centering
    \caption{Effect of the pre-trained text Model (\ie, CLIP) on CrossTask dataset for prediction horizon $T\in\{3,4\}$. SR and mIoU indicate Success Rate (SR) and mean Accuracy (mAcc), respectively. }
    \resizebox{0.47\textwidth}{!}{
    \begin{tabular}{c|cc|cc}
    \toprule
    \multirow{2}{*}{Model}  & \multicolumn{2}{c}{$T$ = 3} & \multicolumn{2}{c}{$T$ = 4} \\
   \cline{2-5}    & SR$\uparrow$ & mAcc$\uparrow$ & SR$\uparrow$ & mAcc$\uparrow$  \\
    \hline
    Full         & 26.40 & 53.02  & 16.49 & 48.00 \\
    w/o CLIP           &  25.83   &  52.39   &  14.70  &  46.18   \\
    w/o CLIP \& w/o EPG    &  24.67    &   49.88    &   13.93    & 44.33\\
    \bottomrule
    \end{tabular}
    }
    \label{tab:ClipAndEventAblate}
    \vspace{-3mm}
\end{table}

\begin{figure*}[tbp]
	\centering
        \vspace{-4mm}		
         \begin{minipage}{0.27\linewidth}
		\centering
            \includegraphics[width=1\linewidth]{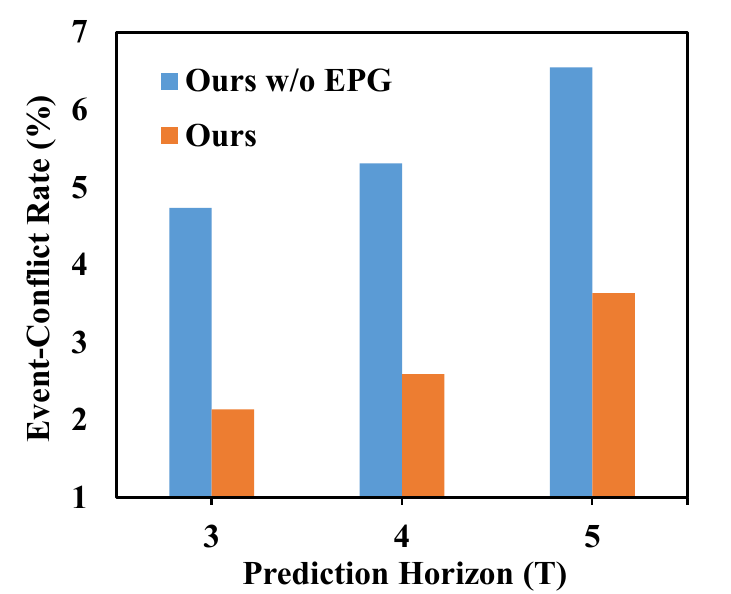}
		\caption{Analysis by event-conflict rate (\ie, the proportion of event-conflict procedures) on the CrossTask dataset.}
		\label{fig:eventConflict}
	\end{minipage}
     \hspace{1em}
	\begin{minipage}{0.7\linewidth}
 \vspace{3mm}
		\centering
		\includegraphics[width=1\linewidth]{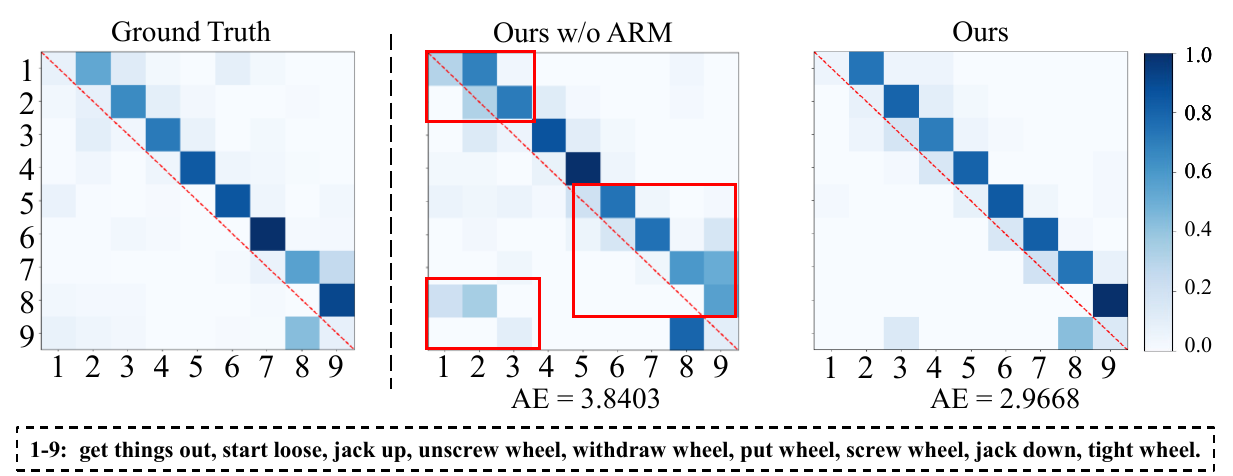}
        \vspace{-6mm}	
	\caption{Visualizations of the action transition matrix for the event ``Change a Tire'' on the CrossTask dataset. The \textit{i-row-j-column} depicts the probability of the transition from \textit{i}-th action to \textit{j}-th action and there are nine actions in total. Darker color indicates higher probability. Best viewed in color.}
		\label{fig:relation_visualization}
	\end{minipage}
        \vspace{-3mm}
\end{figure*}

\textbf{Effect of the Pre-trained Text Model.} 
In our method, we use the pre-trained CLIP for the text representation extraction. 
For a fair comparison, we report the result of a variant of our E3P. 
Specifically, we replace the hand-crafted text prompt (\ie, based on CLIP) with learnable tokens (the same as P3IV~\cite{zhao2022p3iv}) and use one-hot action and order labels as supervision. 
As shown in Table~\ref{tab:ClipAndEventAblate}, even if without the CLIP (\ie, w/o CLIP), our method still achieves the state-of-the-art performance, \ie, 25.83\% when $T=3$ and 14.70\% when $T=4$ in terms of Success Rate (SR). 
Furthermore, we remove the Event-aware Prompt Generator from the above variant (\ie, w/o CLIP \& w/o EPG), our method still outperforms the latest state-of-the-art P3IV~\cite{zhao2022p3iv} (\ie, 1.33\% and 0.53\% on SR of $T=3$ and $T=4$). 
These results demonstrate the effectiveness of our approach again.

\subsection{More Analysis}
\textbf{Quantitative Analysis by Event-Conflict Rate.}~Here, we verify the effectiveness of our Event-aware Prompt Generator (EPG) by calculating an event-conflict rate, \ie, the proportion of event-conflict procedure in all predicted procedures. 
An event-conflict procedure is defined as a procedure with actions that belong to different events. 
As shown in Figure~\ref{fig:eventConflict}, our full model predicts procedures with fewer event-conflicts, compared to our model without the Event-aware Prompt Generator (\ie, Ours w/o EPG). 
These results demonstrate our event-guided paradigm helps exclude some impossible actions in transforming a start state to the goal state, bridging the semantic gap.

\textbf{Analysis by the Action Transition Matrix.}
To verify the effectiveness of our Action Relation Mining (ARM) module, we compute the ground truth action transition matrix and the action transition matrix learned by a model.  
The $i$-row-$j$-column element in the transition matrix depicts the probability of the transition from $i$-th action to $j$-th action (\ie, two successive actions). 
For an intuitive comparison, we first focus on the action transition within an event ``Change a Tire'' and visualize the transition matrix. 
As shown in Figure~\ref{fig:relation_visualization}, the action transition matrix learned by our full model is more consistent with the Ground Truth, compared with our E3P without Action Relation Mining (``Ours w/o ARM''). 
The results demonstrate the effectiveness of our proposed ``mask-and-predict'' approach for relation mining.

For a quantitative analysis, we introduce a quantitative metric, namely Absolute Error (AE), to measure the difference between the learned transition matrix and the corresponding ground truth. 
The AE is defined as $AE = \sum_{i=1}^{n}\sum_{j=1}^{n}  | F_{ij} - E_{ij} | $, where $n$ is the number of actions, $F$ is the learned matrix and $E$ is the ground truth matrix. 
As shown in Figure~\ref{fig:relation_visualization}, our full model achieves a lower AE (2.96) compared with ``Ours w/o ARM'' (3.84) in the ``Change a Tire'' event. 
Furthermore, we calculate the mean Absolute Error (mAE) to measure the difference of action matrices for all events. 
As shown in Table~\ref{tab:relationQuantitative}, our full model achieves a much lower mAE compared with ``Ours w/o ARM'', 
which attributes to our proposed relation mining scheme.

\begin{table}[t]
    \centering
    \caption{Quantitative analysis of the learned transition matrix on CrossTask dataset in terms of mean Absolute Error (mAE) for prediction horizon $T=4$.}
    \begin{tabular}{c|ccc}
    \toprule
      Method   &  Ours & Ours w/o ARM & P3IV~\cite{zhao2022p3iv} \\
      \hline
      mAE ↓ &    2.55  &  3.21    &    3.56  \\
    \bottomrule
    \end{tabular}
    \label{tab:relationQuantitative}
\end{table}

\begin{figure}[t]
    \centering
    \includegraphics[width=0.47\textwidth]{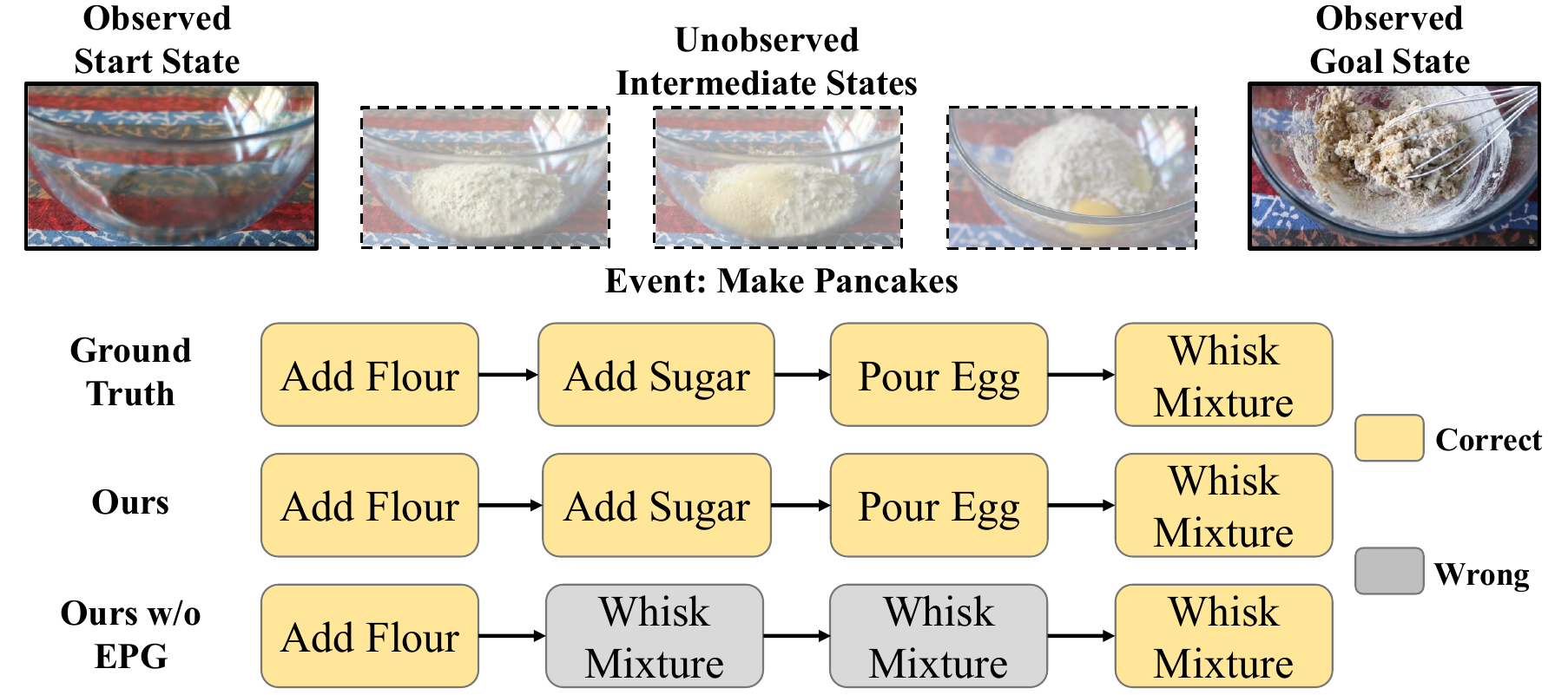}
    \caption{Quantitative analysis of the Event-aware Prompt Generator (EPG). In this figure, we take a four-action procedure planning as an example. Best viewed in color.} 
    \label{fig:Quarlitative_EAP}
\end{figure}

\begin{figure}[t]
    \centering
    \includegraphics[width=0.47\textwidth]{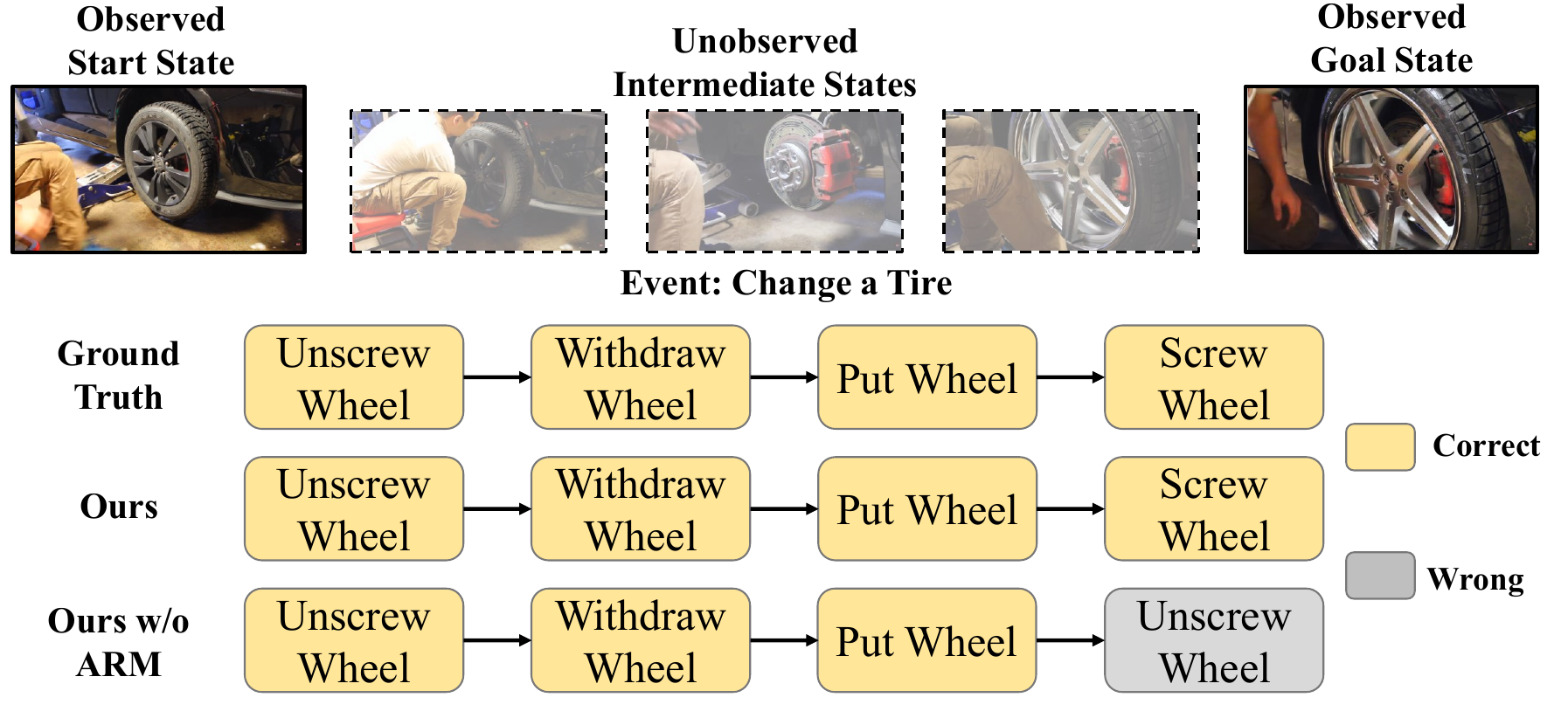}
    \caption{Quantitative analysis of the Action Relation Mining (ARM) module. In this figure, we take a four-action procedure as an example. Best viewed in color.}  
    \label{fig:Quarlitative_IARM}
\end{figure}

\textbf{Qualitative analysis of the Event-aware Prompt Generator (EPG).} 
In Figure~\ref{fig:Quarlitative_EAP}, we give an example to demonstrate the effect of our EPG. 
Due to the semantic gap between the observed start-goal states and unobserved intermediate actions, it is hard to predict ``Add Sugar'' and ``Pour Egg'' without Event-aware Prompt Generator, as shown by ``Ours w/o EPG''. 
In contrast, our full model captures the necessity of adding some sugar and eggs to make a fluffy pancake, which shows the effectiveness of our event-guided paradigm.

\textbf{Qualitative Analysis of the Action Relation Mining.}
In Figure~\ref{fig:Quarlitative_IARM}, we give an example to demonstrate the effect of our ARM. 
We find that ``Ours w/o ARM'' incorrectly predicts the ``Screw Wheel'' as ``Unscrew Wheel'', since it does not capture the action association that ``Screw Wheel'' should follow ``Put Wheel''. 
In contrast, our full model predicts the correct procedure, which shows the importance of modeling action relations and the effectiveness of our proposed ARM.

\section{Conclusion}
In this work, we propose a novel event-guided paradigm to bridge the semantic gap between the observed visual states and unobserved intermediate actions, aiming at solving procedure planning from instructional videos with text supervision.
Based on the paradigm, we proposed an Event-guided Prompting-based Procedure Planning (E3P) model, which encodes event information into the sequential modeling to support procedure planning. 
A mask-and-predict approach  is adopted to fully consider the strong action associations within each event. 
Extensive experiments on three datasets demonstrate the effectiveness of our event-guided paradigm, and our E3P achieves a new state-of-the-art performance. 
One limitation of both previous works and our work is that, they perform poorly when being evaluated on events that do not belong to the training set (\ie, Success Rate $< 1\%$), the future effort could be devoted to such cross-event procedure planning from instructional videos. 

\small{\textbf{Acknowledgement. }
This work was supported partially by the NSFC (U21A20471,U1911401,U1811461,62206315), Guangdong NSF Project (No. 2023B1515040025, 2020B1515120085), Fundamental Research Funds for the Central Universities, SYSU (23ptpy112), China Postdoctoral SF (2022M713574)
}

{\small
\bibliographystyle{IEEEtran}
\bibliography{reference}

\begin{thebibliography}{10}\itemsep=-1pt

\bibitem{abu2018will}
Yazan Abu~Farha, Alexander Richard, and Juergen Gall.
\newblock When will you do what?-anticipating temporal occurrences of
  activities.
\newblock In {\em CVPR}, 2018.

\bibitem{alayrac2016NIV}
Jean-Baptiste Alayrac, Piotr Bojanowski, Nishant Agrawal, Josef Sivic, Ivan
  Laptev, and Simon Lacoste-Julien.
\newblock Unsupervised learning from narrated instruction videos.
\newblock In {\em CVPR}, 2016.

\bibitem{andrychowicz2020Robotic}
OpenAI:~Marcin Andrychowicz, Bowen Baker, Maciek Chociej, Rafal Jozefowicz, Bob
  McGrew, Jakub Pachocki, Arthur Petron, Matthias Plappert, Glenn Powell, Alex
  Ray, et~al.
\newblock Learning dexterous in-hand manipulation.
\newblock {\em IJRR}, 2020.

\bibitem{ba2013adaptiveDrop}
Jimmy Ba and Brendan Frey.
\newblock Adaptive dropout for training deep neural networks.
\newblock {\em NeurIPS}, 2013.

\bibitem{2021EXTGAIL}
Jing {Bi}, Jiebo {Luo}, and Chenliang {Xu}.
\newblock Procedure planning in instructional videos via contextual modeling
  and model-based policy learning.
\newblock In {\em ICCV}, 2021.

\bibitem{carreira2017deep3D1}
Joao Carreira and Andrew Zisserman.
\newblock Quo vadis, action recognition? a new model and the kinetics dataset.
\newblock In {\em CVPR}, 2017.

\bibitem{chang2020procedure}
Chien-Yi Chang, De-An Huang, Danfei Xu, Ehsan Adeli, Li Fei-Fei, and
  Juan~Carlos Niebles.
\newblock Procedure planning in instructional videos.
\newblock In {\em ECCV}, 2020.

\bibitem{chung2014GRN}
Junyoung Chung, Caglar Gulcehre, KyungHyun Cho, and Yoshua Bengio.
\newblock Empirical evaluation of gated recurrent neural networks on sequence
  modeling.
\newblock {\em arXiv}, 2014.

\bibitem{donahue2015RnnsActionRecognition1}
Jeffrey Donahue, Lisa Anne~Hendricks, Sergio Guadarrama, Marcus Rohrbach,
  Subhashini Venugopalan, Kate Saenko, and Trevor Darrell.
\newblock Long-term recurrent convolutional networks for visual recognition and
  description.
\newblock In {\em CVPR}, 2015.

\bibitem{dosovitskiy2020VIT1}
Alexey Dosovitskiy, Lucas Beyer, Alexander Kolesnikov, Dirk Weissenborn,
  Xiaohua Zhai, Thomas Unterthiner, Mostafa Dehghani, Matthias Minderer, Georg
  Heigold, Sylvain Gelly, et~al.
\newblock An image is worth 16x16 words: Transformers for image recognition at
  scale.
\newblock {\em arXiv}, 2020.

\bibitem{ehsani2018let}
Kiana Ehsani, Hessam Bagherinezhad, Joseph Redmon, Roozbeh Mottaghi, and Ali
  Farhadi.
\newblock Who let the dogs out? modeling dog behavior from visual data.
\newblock In {\em CVPR}, 2018.

\bibitem{2019UAAA}
Yazan~Abu Farha and Juergen Gall.
\newblock Uncertainty-aware anticipation of activities.
\newblock In {\em ICCV}, 2019.

\bibitem{girdhar2019videoTransformer1}
Rohit Girdhar, Joao Carreira, Carl Doersch, and Andrew Zisserman.
\newblock Video action transformer network.
\newblock In {\em CVPR}, 2019.

\bibitem{gong2022future}
Dayoung Gong, Joonseok Lee, Manjin Kim, Seong~Jong Ha, and Minsu Cho.
\newblock Future transformer for long-term action anticipation.
\newblock In {\em CVPR}, 2022.

\bibitem{graves2012LSTM}
Alex Graves and Alex Graves.
\newblock Long short-term memory.
\newblock {\em Supervised sequence labelling with recurrent neural networks},
  2012.

\bibitem{gutmann2010contrastive}
Michael Gutmann and Aapo Hyv{\"a}rinen.
\newblock Noise-contrastive estimation: A new estimation principle for
  unnormalized statistical models.
\newblock In {\em AISTATS}, 2010.

\bibitem{he2016deep2Dconv3}
Kaiming He, Xiangyu Zhang, Shaoqing Ren, and Jian Sun.
\newblock Deep residual learning for image recognition.
\newblock In {\em CVPR}, 2016.

\bibitem{hinton2012dropout}
Geoffrey~E Hinton, Nitish Srivastava, Alex Krizhevsky, Ilya Sutskever, and
  Ruslan~R Salakhutdinov.
\newblock Improving neural networks by preventing co-adaptation of feature
  detectors.
\newblock {\em arXiv}, 2012.

\bibitem{ju2022prompting}
Chen Ju, Tengda Han, Kunhao Zheng, Ya Zhang, and Weidi Xie.
\newblock Prompting visual-language models for efficient video understanding.
\newblock In {\em ECCV}, 2022.

\bibitem{kataoka2016shorttermanticipation1}
Hirokatsu Kataoka, Yudai Miyashita, Masaki Hayashi, Kenji Iwata, and Yutaka
  Satoh.
\newblock Recognition of transitional action for short-term action prediction
  using discriminative temporal cnn feature.
\newblock In {\em BMVC}, 2016.

\bibitem{ke2019time}
Qiuhong Ke, Mario Fritz, and Bernt Schiele.
\newblock Time-conditioned action anticipation in one shot.
\newblock In {\em CVPR}, 2019.

\bibitem{kingma2014adam}
Diederik~P Kingma and Jimmy Ba.
\newblock Adam: A method for stochastic optimization.
\newblock {\em arXiv}, 2014.

\bibitem{li2022bridge}
Muheng Li, Lei Chen, Yueqi Duan, Zhilan Hu, Jianjiang Feng, Jie Zhou, and Jiwen
  Lu.
\newblock Bridge-prompt: Towards ordinal action understanding in instructional
  videos.
\newblock In {\em CVPR}, 2022.

\bibitem{lin2019deep2Dconv4}
Ji Lin, Chuang Gan, and Song Han.
\newblock Tsm: Temporal shift module for efficient video understanding.
\newblock In {\em ICCV}, 2019.

\bibitem{liu2023pre}
Pengfei Liu, Weizhe Yuan, Jinlan Fu, Zhengbao Jiang, Hiroaki Hayashi, and
  Graham Neubig.
\newblock Pre-train, prompt, and predict: A systematic survey of prompting
  methods in natural language processing.
\newblock {\em ACM Computing Surveys}, 2023.

\bibitem{liu2021VIT2}
Ze Liu, Yutong Lin, Yue Cao, Han Hu, Yixuan Wei, Zheng Zhang, Stephen Lin, and
  Baining Guo.
\newblock Swin transformer: Hierarchical vision transformer using shifted
  windows.
\newblock In {\em ICCV}, 2021.

\bibitem{medsker2001RNN}
Larry~R Medsker and LC Jain.
\newblock Recurrent neural networks.
\newblock {\em Design and Applications}, 2001.

\bibitem{miech2020s3d}
Antoine Miech, Jean-Baptiste Alayrac, Lucas Smaira, Ivan Laptev, Josef Sivic,
  and Andrew Zisserman.
\newblock End-to-end learning of visual representations from uncurated
  instructional videos.
\newblock In {\em CVPR}, 2020.

\bibitem{miech2019howto100m}
Antoine Miech, Dimitri Zhukov, Jean-Baptiste Alayrac, Makarand Tapaswi, Ivan
  Laptev, and Josef Sivic.
\newblock Howto100m: Learning a text-video embedding by watching hundred
  million narrated video clips.
\newblock In {\em ICCV}, 2019.

\bibitem{nag2022zero}
Sauradip Nag, Xiatian Zhu, Yi-Zhe Song, and Tao Xiang.
\newblock Zero-shot temporal action detection via vision-language prompting.
\newblock In {\em ECCV}, 2022.

\bibitem{neimark2021videoTransformer2}
Daniel Neimark, Omri Bar, Maya Zohar, and Dotan Asselmann.
\newblock Video transformer network.
\newblock In {\em ICCV}, 2021.

\bibitem{paszke2017pytorch}
Adam Paszke, Sam Gross, Soumith Chintala, Gregory Chanan, Edward Yang, Zachary
  DeVito, Zeming Lin, Alban Desmaison, Luca Antiga, and Adam Lerer.
\newblock Automatic differentiation in pytorch.
\newblock 2017.

\bibitem{radford2021CLIP}
Alec Radford, Jong~Wook Kim, Chris Hallacy, Aditya Ramesh, Gabriel Goh,
  Sandhini Agarwal, Girish Sastry, Amanda Askell, Pamela Mishkin, Jack Clark,
  et~al.
\newblock Learning transferable visual models from natural language
  supervision.
\newblock In {\em ICML}, 2021.

\bibitem{schick2020exploiting}
Timo Schick and Hinrich Sch{\"u}tze.
\newblock Exploiting cloze questions for few shot text classification and
  natural language inference.
\newblock {\em arXiv}, 2020.

\bibitem{shao2020deep2Dconv5}
Hao Shao, Shengju Qian, and Yu Liu.
\newblock Temporal interlacing network.
\newblock In {\em AAAI}, 2020.

\bibitem{simonyan2014tsn}
Karen Simonyan and Andrew Zisserman.
\newblock Two-stream convolutional networks for action recognition in videos.
\newblock {\em NeurIPS}, 2014.

\bibitem{2018UPN}
Aravind Srinivas, Allan Jabri, Pieter Abbeel, Sergey Levine, and Chelsea Finn.
\newblock Universal planning networks: Learning generalizable representations
  for visuomotor control.
\newblock In {\em ICML}, 2018.

\bibitem{sun2022plate}
Jiankai Sun, De-An Huang, Bo Lu, Yun-Hui Liu, Bolei Zhou, and Animesh Garg.
\newblock Plate: Visually-grounded planning with transformers in procedural
  tasks.
\newblock {\em RA-L}, 2022.

\bibitem{tran2015deep3D2}
Du Tran, Lubomir Bourdev, Rob Fergus, Lorenzo Torresani, and Manohar Paluri.
\newblock Learning spatiotemporal features with 3d convolutional networks.
\newblock In {\em ICCV}, 2015.

\bibitem{ullah2017RnnsActionRecognition3}
Amin Ullah, Jamil Ahmad, Khan Muhammad, Muhammad Sajjad, and Sung~Wook Baik.
\newblock Action recognition in video sequences using deep bi-directional lstm
  with cnn features.
\newblock {\em IEEE access}, 2017.

\bibitem{2017Attention}
A. Vaswani, N. Shazeer, N. Parmar, J. Uszkoreit, L. Jones, A.~N. Gomez, L.
  Kaiser, and I. Polosukhin.
\newblock Attention is all you need.
\newblock {\em arXiv}, 2017.

\bibitem{vaswani2017attentionisallyouneed}
Ashish Vaswani, Noam Shazeer, Niki Parmar, Jakob Uszkoreit, Llion Jones,
  Aidan~N Gomez, {\L}ukasz Kaiser, and Illia Polosukhin.
\newblock Attention is all you need.
\newblock {\em NeurIPS}, 2017.

\bibitem{viterbi1967viterbi}
Andrew Viterbi.
\newblock Error bounds for convolutional codes and an asymptotically optimum
  decoding algorithm.
\newblock {\em TIT}, 1967.

\bibitem{vondrick2016shorttermanticipation2}
Carl Vondrick, Hamed Pirsiavash, and Antonio Torralba.
\newblock Anticipating visual representations from unlabeled video.
\newblock In {\em CVPR}, 2016.

\bibitem{wan2013dropconnect}
Li Wan, Matthew Zeiler, Sixin Zhang, Yann Le~Cun, and Rob Fergus.
\newblock Regularization of neural networks using dropconnect.
\newblock In {\em ICML}, 2013.

\bibitem{wang2021actionclip}
Mengmeng Wang, Jiazheng Xing, and Yong Liu.
\newblock Actionclip: A new paradigm for video action recognition.
\newblock {\em arXiv}, 2021.

\bibitem{wang2013fastDrop}
Sida Wang and Christopher Manning.
\newblock Fast dropout training.
\newblock In {\em ICML}, 2013.

\bibitem{wulfmeier2016autopilot}
Markus Wulfmeier, Dominic~Zeng Wang, and Ingmar Posner.
\newblock Watch this: Scalable cost-function learning for path planning in
  urban environments.
\newblock In {\em IROS}, 2016.

\bibitem{xie2017deep3D3}
Saining Xie, Chen Sun, Jonathan Huang, Zhuowen Tu, and Kevin Murphy.
\newblock Rethinking spatiotemporal feature learning for video understanding.
\newblock {\em arXiv}, 2017.

\bibitem{yue2015RnnsActionRecognition2}
Joe Yue-Hei~Ng, Matthew Hausknecht, Sudheendra Vijayanarasimhan, Oriol Vinyals,
  Rajat Monga, and George Toderici.
\newblock Beyond short snippets: Deep networks for video classification.
\newblock In {\em CVPR}, 2015.

\bibitem{zhang2021VideoTransformer3}
Hao Zhang, Yanbin Hao, and Chong-Wah Ngo.
\newblock Token shift transformer for video classification.
\newblock In {\em ACM MM}, 2021.

\bibitem{zhang2021videoTransformer4}
Yanyi Zhang, Xinyu Li, Chunhui Liu, Bing Shuai, Yi Zhu, Biagio Brattoli, Hao
  Chen, Ivan Marsic, and Joseph Tighe.
\newblock Vidtr: Video transformer without convolutions.
\newblock In {\em ICCV}, 2021.

\bibitem{zhao2022p3iv}
He Zhao, Isma Hadji, Nikita Dvornik, Konstantinos~G Derpanis, Richard~P Wildes,
  and Allan~D Jepson.
\newblock P3iv: Probabilistic procedure planning from instructional videos with
  weak supervision.
\newblock In {\em CVPR}, 2022.

\bibitem{zhukov2019crosstask}
Dimitri Zhukov, Jean-Baptiste Alayrac, Ramazan~Gokberk Cinbis, David Fouhey,
  Ivan Laptev, and Josef Sivic.
\newblock Cross-task weakly supervised learning from instructional videos.
\newblock In {\em CVPR}, 2019.

\end{thebibliography}
}

\end{document}


\renewcommand{\thetable}{A\arabic{table}}
\renewcommand{\thefigure}{A\arabic{figure}}
\renewcommand\thesection{\Alph{section}}
\renewcommand{\citeform}[1]{A#1} \makeatletter \def\@biblabel#1{[A#1]} \makeatother
\renewcommand{\thefootnote}{}

\title{Appendix \emph{for}

Event-Guided Procedure Planning from Instructional Videos with Text Supervision}

\author{An-Lan Wang$^{*}$, Kun-Yu Lin$^{*}$, Jia-Run Du, Jingke Meng$^{{\dagger}}$, Wei-Shi Zheng$^{{\dagger}}$\\
\normalsize School of Computer Science and Engineering, Sun Yat-sen University, China\\
\normalsize Key Laboratory of Machine Intelligence and Advanced Computing, Ministry of Education, China\\
{\tt\small \{wanganlan, linky5, dujr6\}@mail2.sysu.edu.cn, mengjke@gmail.com, wszheng@ieee.org}
}

\maketitle
\ificcvfinal\thispagestyle{empty}\fi

\footnotetext{ * indicates equal contribution. $\dagger$ indicates the corresponding author.}

\section{Appendix Overview}
Due to the lack of space in the main manuscript, we provide more specific details of our Event-guided Prompting-based Procedure Planning (E3P) in the \textit{Appendix}, organized as follows: Section~\ref{implementationDetail} provides a more detailed description of our E3P implementation. Section~\ref{additionalExperiments} includes several additional experiments.

\section{Implementation Details}
\label{implementationDetail}
Following previous work~\cite{zhao2022p3iv}, we use pre-extracted visual and text features. The dimension of the pre-extracted visual and text features is 512, we use two multi-layer perceptrons (MLP) with shape $[512\rightarrow256\rightarrow128]$ interspersed with ReLU to embed the original visual and text feature, respectively. For the Event-aware Prompt Generator, the event-information extractor is implemented using a MLP with shape $[256\rightarrow64\rightarrow128]$, and the event-information aggregator is a Transformer encoder of one self-attention layer with 128-dimensional hidden states. For the Action Relation Mining module, we use two masked self-attention layers followed by a feed-forward network (FFN).

\section{Additional Experiments}
\label{additionalExperiments}
In addition to the various ablations reported in the main manuscript, we provide some additional experiments to verify the effectiveness of our proposed Event-guided Prompting-based Procedure Planning (E3P).

\subsection{Effect of the Event-information Aggregator}
\label{EIAAblation}
For the event-information aggregator, we provide two implementations, \textit{\textbf{Concat}}, \textit{\textbf{Transf}} (\ie, used in the main manuscript):
\begin{itemize}
    \item \textit{\textbf{Concat}} first concatenates the prompt representation with the event information and then uses a MLP to project to its original dimension. 
    \item \textit{\textbf{Transf}} means using a Transformer encoder of one self-attention layer to process the $T+1$ Tokens, \ie, $T$ prompt representations and one event information token.
\end{itemize}

In Table \ref{tab:EPGAblate}, we conduct experiments using different event-information aggregator implementations. ``\textit{Transf}'' outperforms ``\textit{Concat}'' in all prediction horizon $T\in\{3,4\}$ (\ie, 0.63\% when $T=3$ and 1.08\% when $T=4$ in terms of Success Rate). In addition, ``\textit{concat}'' still achieves state-of-the-art performance, which demonstrates the effectiveness of the proposed event-guided paradigm.

\begin{table}[t]
    \centering
    \caption{Effect of Event-information Aggregator for prediction horizon $T\in\{3,4\}$ on CrossTask dataset. SR and mAcc indicate Success Rate and mean Accuracy, respectively. P3IV is the latest state-of-the-art method. }
    \begin{tabular}{c|cc|cc}
    \toprule
         \multirow{2}{*}{Model}  & \multicolumn{2}{c}{$T$ = 3} & \multicolumn{2}{c}{$T$ = 4} \\
   \cline{2-5}    & SR$\uparrow$ & mAcc$\uparrow$ & SR$\uparrow$ & mAcc$\uparrow$  \\
    \hline
     Concat  & 25.77 & 51.32 & 15.41 & 47.44\\
     Transf & 26.40 & 53.02 & 16.49 & 48.00\\
     \hline
     P3IV~\cite{zhao2022p3iv} & 23.34 & 49.96 & 13.40 & 44.16 \\
    \bottomrule
    \end{tabular}
    
    \label{tab:EPGAblate}
\end{table}

\begin{table}[t]
    \centering
    \caption{Quantitative analysis of the number of masked self-attention layers used in the Action Relation Mining module for prediction $T\in\{3,4\}$ on CrossTask dataset. SR and mAcc indicate Success Rate and mean Accuracy, respectively.}
    \begin{tabular}{c|cc|cc}
    \toprule
         \multirow{2}{*}{Number of Layers}  & \multicolumn{2}{c}{$T$ = 3} & \multicolumn{2}{c}{$T$ = 4} \\
   \cline{2-5}    & SR$\uparrow$ & mAcc$\uparrow$ & SR$\uparrow$ & mAcc$\uparrow$  \\
    \hline
        1 & 25.97  &  52.69 & 16.10  & 47.50  \\
        2 & 26.26  &  52.91 & 16.49  & 48.00  \\
        3 & 26.14  &  52.77 & 16.15  & 47.69  \\
        4 & 25.56  &  52.42 & 15.69  & 47.43  \\
        \bottomrule
    \end{tabular}
    \label{tab:ARMLayerAblation}
\end{table}

\subsection{Analysis of the number of layers used in the Action Relation Mining}
\label{Section:ARMLayerablation}
In Table \ref{tab:ARMLayerAblation}, we ablate the number of masked self-attention layers used in the Action Relation Mining module (drop rate is 0.2). The results show that using two masked self-attention layers (\ie. used in the main manuscript) attains the best performance, \ie, 26.26\% when $T=3$ and 16.49\% when $T=4$ in terms of Success Rate (SR).

\begin{table}[t]
    \centering
    \caption{Comparison to previous state-of-the-art methods using Visual State Supervision for prediction $T=3$ on CrossTask dataset, in terms of Success Rate (SR), mean Accuracy (mAcc), and mean Intersection over Union (mIoU).}
    \begin{tabular}{c|ccc}
    \toprule
        Methods & SR$\uparrow$ & mAcc$\uparrow$ & mIoU$\uparrow$  \\
    \hline
        baseline & 22.86 & 47.87 & 70.34\\
        + event-guided paradigm & 25.70 & 53.19 & 72.76   \\
    \hline
        P3IV~\cite{zhao2022p3iv} with visual sup & 24.41 & 45.17 & 73.83\\
    \bottomrule
    \end{tabular}
    \label{tab:strongSup}
\end{table}

\subsection{Effect of the event-guide paradigm}
To verify the effect of the event-guided paradigm in Procedure Planning from instructional videos with Visual Supervision (PPVS), we conduct an experiment that adopts the event-guided paradigm to a variant of P3IV~\cite{zhao2022p3iv}. In this variant (\ie, baseline), we remove the adversarial strategy and use intermediate visual states as supervision. Then, we insert our proposed Event-guided Prompt Generator (EPG) into this variant (\ie, + event-guided paradigm), but instead of hand-craft prompts, the input to this EPG is learnable queries. The results are shown in \ref{tab:strongSup}, by introducing the event-guided paradigm, we attain a significant improvement (\eg, 1.84\% in terms of Success Rate), outperforming P3IV\cite{zhao2022p3iv} with visual state supervision (\ie, P3IV with visual sup). These consistent results demonstrate the effectiveness of our proposed event-guided paradigm for PPVS.

{\small
\bibliographystyle{ieee_fullname}
\bibliography{egbib}
}